\documentclass[final,5p,times,twocolumn]{elsarticle}
\usepackage{amssymb}
\usepackage{amsmath}
\usepackage{amsthm}
\usepackage{algorithm}
\usepackage{algpseudocode}  
\usepackage{url}            
\usepackage{booktabs}       
\usepackage{nicefrac}       
\usepackage{microtype}      
\usepackage[table,xcdraw]{xcolor}         
\usepackage{multirow}
\usepackage{graphicx}
\usepackage{subcaption}
\usepackage{pifont}
\usepackage{hyperref}       

\newtheorem{proposition}{Proposition}

\journal{Knowledge-Based Systems}

\begin{document}

\begin{frontmatter}

\title{HACD: Heterogeneous Asymmetric Contrastive Distillation}
\author[hosei]{Qi Wang}
\ead{qi.wang.7c@stu.hosei.ac.jp}

\author[hosei]{Jinjia Zhou\corref{cor1}}
\ead{zhou@hosei.ac.jp}

\cortext[cor1]{Corresponding author.}

\affiliation[hosei]{organization={Graduate School of Science and Engineering, Hosei University},
addressline={3-7-2 Kajino-cho, Koganei-shi},
city={Tokyo},
postcode={184-8584},
country={Japan}}

\begin{abstract}
Knowledge distillation (KD) between architectural families, e.g., from a Transformer teacher to a CNN student, is difficult because the two networks learn different representations. Logits share a label space, but heterogeneous teachers differ in logit offset, scale, and sharpness, and their non-target classes are less semantically organized. Conventional logit distillation matches these values sample by sample, asking the student to imitate how another architecture expresses its predictions. We argue that a heterogeneous teacher should pass on the relational structure of its predictions, i.e., how it arranges samples relative to one another, and propose Heterogeneous Asymmetric Contrastive Distillation (HACD) to transfer this structure. HACD maps teacher and student logits to centered, temperature-softened probability directions on a shared unit hypersphere, which removes architecture-dependent offsets and lets the temperature control how sharply peaked outputs enter the representation while keeping informative non-target classes. Each student output must then identify its own teacher target within the batch. Because every teacher target also acts as a negative for the others, HACD computes the targets on a moderately augmented view and contrasts both a moderate and a strong student view against them. HACD adds no learnable parameters and needs one teacher pass per sample. It achieves the best or second-best accuracy on 11 of 12 heterogeneous pairs on CIFAR-100, without ground-truth labels, and on 14 of 15 pairs on ImageNet-1K, improving over the reported vanilla KD results by 8.82 and 1.33 percentage points on average. Checkpoints are available at \url{https://github.com/BTAEYANG/HACD-Heterogeneous-Asymmetric-Contrastive-Distillation}; the code will be released upon publication.
\end{abstract}

\end{frontmatter}

\section{Introduction}
\label{sec:intro}
Vision Transformers and MLP-based models achieve recognition performance competitive with convolutional neural networks (CNNs) \citep{vit,swin,mlpmixer}, but their size and computational cost hinder deployment on resource-constrained devices. Knowledge distillation (KD)~\citep{kd} addresses this problem by transferring the knowledge of a high-capacity teacher to a compact student and has been studied extensively in visual recognition \citep{gou2021knowledge}. Most KD methods, however, were developed for homogeneous pairs in which both networks are CNNs. Although such a teacher and student may differ in depth, width, or capacity, they share operators, inductive biases, and stage-wise structures, so their intermediate features correspond closely enough to be aligned through layer matching, projection, or spatial transformation, and their logits are produced by classifiers built on similar representations.

\begin{figure*}[t]
\centering
\begin{subfigure}{0.235\textwidth}
\centering
\includegraphics[clip, trim=8 8 8 20, width=\textwidth]{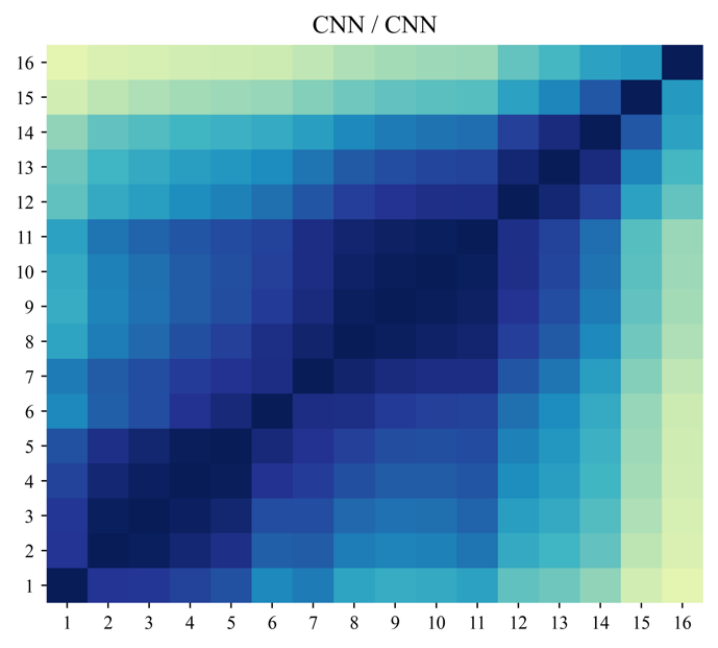}
\caption{CNN-to-CNN}
\label{fig:cnn_cnn}
\end{subfigure}
\hfill
\begin{subfigure}{0.235\textwidth}
\centering
\includegraphics[clip, trim=8 8 8 20, width=\textwidth]{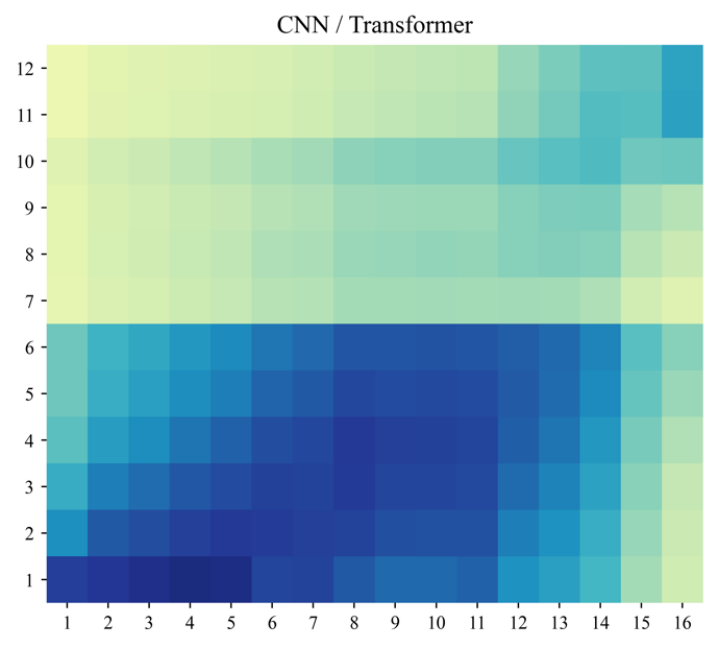}
\caption{CNN-to-Trans.}
\label{fig:cnn_transformer}
\end{subfigure}
\hfill
\begin{subfigure}{0.235\textwidth}
\centering
\includegraphics[clip, trim=8 8 8 20, width=\textwidth]{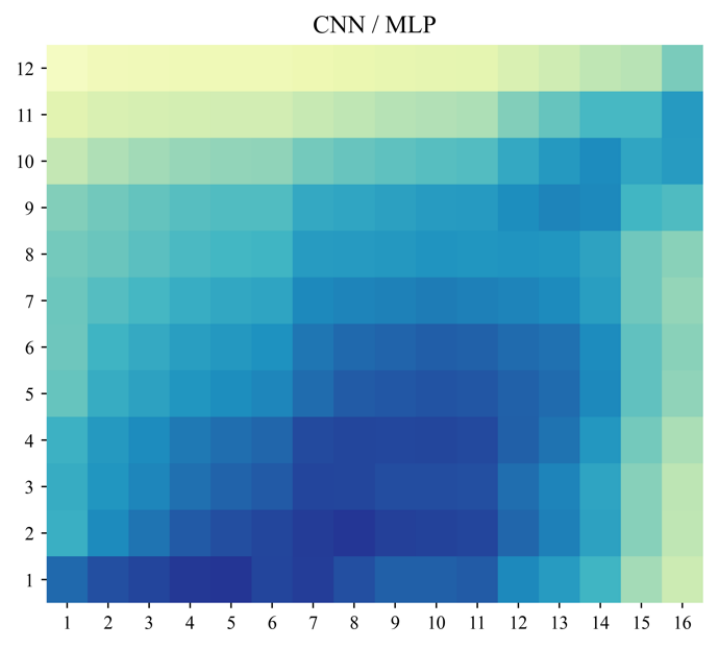}
\caption{CNN-to-MLP}
\label{fig:cnn_mlp}
\end{subfigure}
\hfill
\begin{subfigure}{0.235\textwidth}
\centering
\includegraphics[clip, trim=8 8 8 20, width=\textwidth]{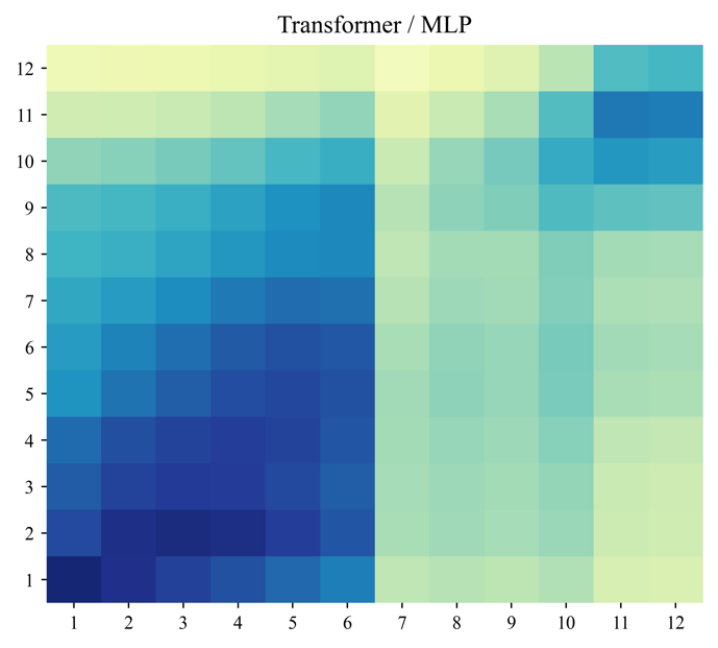}
\caption{Trans.-to-MLP}
\label{fig:transformer_mlp}
\end{subfigure}
\caption{CKA similarity analysis of representations learned by homogeneous and heterogeneous architectures. Each entry shows the CKA similarity between the representations of a layer of the first network (horizontal axis) and a layer of the second network (vertical axis), with darker colors indicating higher similarity. The CNN-to-CNN pair exhibits more consistent representation patterns, whereas heterogeneous pairs involving CNNs, Transformers, and MLPs show substantial discrepancies.}
\label{fig:cka_six_subfigures}
\end{figure*}

Heterogeneous distillation lets a compact CNN student intended for deployment learn from strong Transformer or MLP teachers instead of only from larger CNNs. Feature-based distillation, however, assumes that teacher and student representations can be meaningfully aligned, and this assumption becomes less reliable across architectural families: CNNs organize visual information through spatial feature maps, Transformers model interactions among image tokens, and MLP-based architectures employ distinct token- and channel-mixing mechanisms. In the CKA analysis of Fig.~\ref{fig:cka_six_subfigures}, the layers of two CNNs correspond consistently, whereas pairs that mix CNNs, Transformers, and MLPs do not, so intermediate representations from different families may not be directly comparable even after dimensional or spatial adaptation. Existing heterogeneous methods close this representational gap with projection or adaptation modules \cite{cakd,ofa,pat}, at the cost of additional learnable parameters (Section~\ref{sec:ablation}).

\begin{figure*}[t]
    \centering
    \begin{subfigure}[t]{0.32\textwidth}
        \centering
        \includegraphics[width=\linewidth]{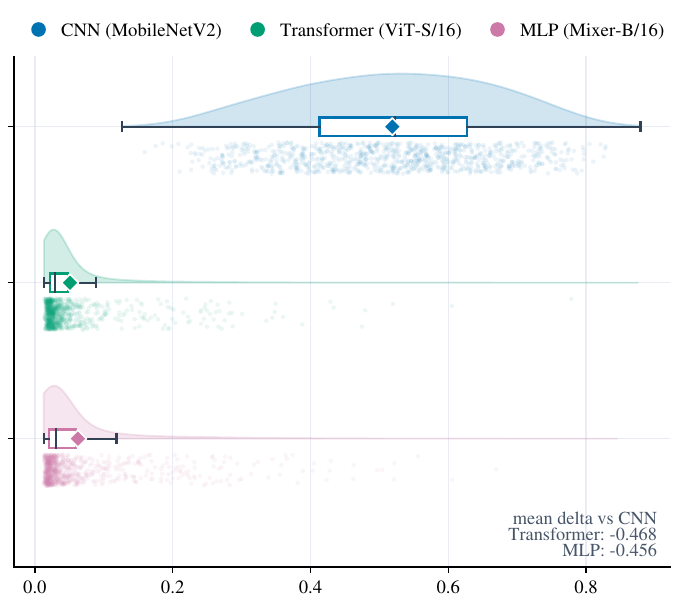}
        \caption{Normalized entropy}
        \label{fig:fig2_entropy}
    \end{subfigure}
    \hfill
    \begin{subfigure}[t]{0.32\textwidth}
        \centering
        \includegraphics[width=\linewidth]{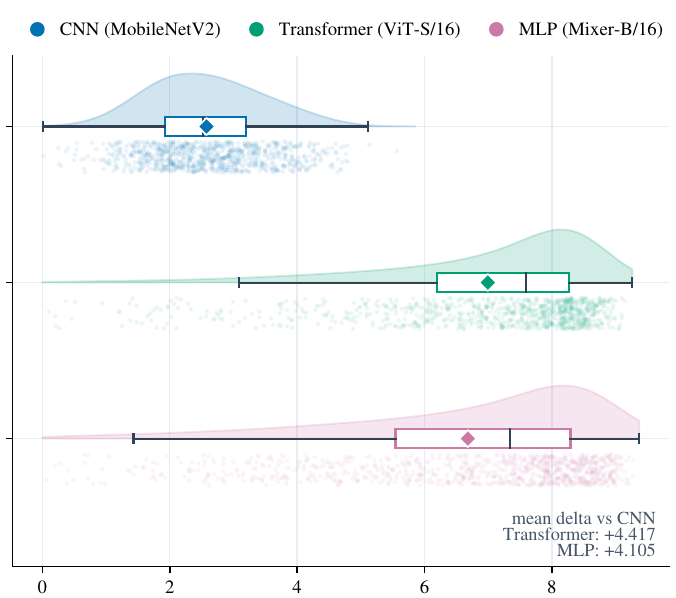}
        \caption{Top-1 / top-2 margin}
        \label{fig:fig2_margin}
    \end{subfigure}
    \hfill
    \begin{subfigure}[t]{0.32\textwidth}
        \centering
        \includegraphics[width=\linewidth]{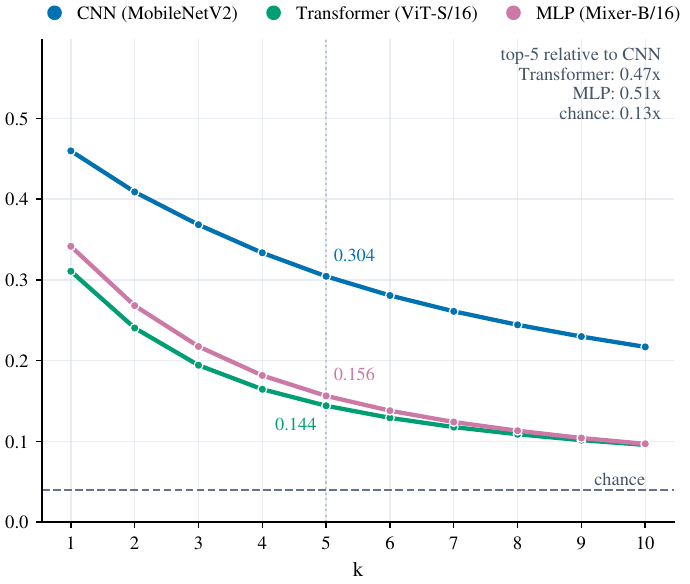}
        \caption{Semantic consistency}
        \label{fig:fig2_semantic}
    \end{subfigure}
    \caption{Output statistics of CNN, Transformer, and MLP teachers on the CIFAR-100 training dataset. Logits are standardized per sample, $\tilde z=(z-\bar z)/\sigma_z$, to remove architecture-dependent scales. (a) Normalized entropy of $\operatorname{softmax}(\tilde z)$. (b) Top-1/top-2 margin of $\tilde z$. (c) Semantic consistency: fraction of the top-$k$ non-target classes that share the superclass of the ground-truth class (dashed line: chance level, $4/99$). In (a) and (b), half violins, boxes, and diamonds show the distribution, quartiles, and mean over samples. The Transformer and MLP teachers produce sharper outputs with less semantically organized non-target classes.}
    \label{fig:logit_statistics}
\end{figure*}

Logits offer a natural alternative, since they share the same label space across architectures and require no feature alignment. Sharing a label space, however, does not make the outputs directly comparable. As shown in Fig.~\ref{fig:logit_statistics}, even after the architecture-dependent logit scale is removed by standardization, the Transformer and MLP teachers produce much sharper distributions than the CNN teacher, with lower normalized entropy (Fig.~\ref{fig:fig2_entropy}) and larger top-1/top-2 margins (Fig.~\ref{fig:fig2_margin}). Their non-target classes, which receive less probability mass under such concentrated predictions \cite{dkd,ranking}, are also less semantically organized: among the five highest-ranked non-target classes, only 14\%--16\% belong to the superclass of the ground-truth class, about half of the 30\% observed for the CNN teacher, although still well above chance level (Fig.~\ref{fig:fig2_semantic}). Since all teachers see the same images, these differences come from the architectures themselves. Conventional KL-based distillation matches the teacher and student distributions independently for each sample \cite{dist}, so it asks a CNN student to reproduce a sharply peaked distribution together with a weakly structured non-target tail, both of which differ from what the student itself produces. Recent logit-based methods improve how these values are matched, by decoupling target and non-target knowledge \cite{dkd}, standardizing logits \cite{lskd}, extracting global and multi-scale local logits \cite{global_local,sdd}, or adopting alternative distribution metrics \cite{wkd}. DIST and MLKD additionally match correlations among the predictions of different samples \cite{dist,mlkd}, but these correlations are still computed from the softened predictions of the teacher, whose sharpness differs across architectures.

We argue that a heterogeneous student can inherit the knowledge of a teacher by arranging samples as the teacher does, rather than by reproducing the values of the teacher alone. A teacher prediction for one image is informative mainly in relation to its predictions for other images: two images of the same class differ in their non-target structure, and an ambiguous image lies closer to some samples than to others. Such relations are defined by comparisons among the outputs of a single network, so transferring them depends less on the offset, scale, and sharpness of the teacher than matching its values does. The relations must also be measured in a suitable space. Relational and contrastive methods such as RKD \cite{relational} and CRD \cite{crd} already transfer inter-sample structure, but they do so in feature space, where the mismatch of Fig.~\ref{fig:cka_six_subfigures} persists; in the heterogeneous setting, their average gains on CIFAR-100 are $-1.40$ and $-0.02$~pp over the students trained from scratch (Table~\ref{tab:cifar100_results}). Heterogeneous distillation therefore needs to transfer relations in a space that both architectures share and in which their outputs are comparable.

Following this view, we propose Heterogeneous Asymmetric Contrastive Distillation (HACD), a logit-only framework built on three design choices. First, to make the relations comparable, HACD maps teacher and student logits to centered, temperature-softened probability directions on a shared unit hypersphere, which removes architecture-dependent logit offsets and normalizes the representation, while the temperature controls how much of the sharpness of a teacher enters it and retains the informative non-target classes of sharply peaked teachers. Second, to transfer the relational structure of the teacher outputs, the student must identify its own teacher target among those of the other samples in the mini-batch through a bidirectional contrast, so that each student output is aligned with its own target and placed relative to the targets of the other samples. Third, the primary teacher targets must be reliable: in a contrastive objective, each teacher target is the positive of one sample and a negative of all others, so unreliable targets, when they serve as the primary supervision, distort the relations of the whole batch. HACD therefore computes the teacher targets on a moderately augmented view, and aligns both a moderate and a strongly augmented student view with these targets, so that the student benefits from strong augmentation while its supervision remains anchored to the moderate view.

HACD achieves the best or second-best accuracy on 11 of the 12 heterogeneous teacher-student pairs on CIFAR-100 and on 14 of the 15 pairs on ImageNet-1K, and improves over the reported results of conventional KD by 8.82 and 1.33 percentage points (pp) on average, respectively. Under the same training configuration, HACD is 5.32 and 1.29~pp more accurate than KD. On CIFAR-100, it uses no ground-truth labels in distillation, so the teacher outputs, compared through the contrastive objective, provide enough supervision on their own. Our main contributions are summarized as follows:

\begin{itemize}
    \item We analyze the logits of CNN, Transformer, and MLP teachers and show that their values carry architecture-specific properties, including sharpness and the organization of non-target classes. Based on this analysis, we rethink heterogeneous logit distillation and design it to transfer the relations among teacher outputs instead of their values.

    \item We propose HACD, a logit-only distillation method that maps teacher and student logits to centered, temperature-softened probability directions on a shared hypersphere and aligns them through bidirectional instance-level contrast, with an analytical characterization of how the temperature balances tail suppression against instance discrimination. The teacher encodes only a moderately augmented view, and its outputs supervise both a moderate and a strong student view, so HACD needs a single teacher forward pass per sample and no additional learnable parameters.

    \item Experiments on 27 heterogeneous teacher-student pairs on CIFAR-100 and ImageNet-1K show that HACD achieves the best or second-best accuracy on 25 pairs, and on CIFAR-100 it does so without ground-truth labels during distillation.
\end{itemize}

\section{Related Work}
\label{related_work}

\subsection{Knowledge Distillation}

Knowledge distillation transfers knowledge from a pretrained teacher to a student, typically through either final predictions or intermediate representations.
\textbf{Logit-based distillation.} Conventional KD minimizes the Kullback-Leibler (KL) divergence between the softened predictions of the teacher and the student \citep{kd}. Subsequent methods refine how this output-level knowledge is used: DKD decouples target-class and non-target-class knowledge \citep{dkd}, and LSKD standardizes logits to remove the influence of their magnitudes \citep{lskd}. Other methods compare class probabilities through the Wasserstein distance \citep{wkd}, preserve the ranking of class predictions \citep{ranking}, or construct multi-scale local logits by pooling the classifier outputs of the final feature map over local regions \citep{sdd}. These methods change which values of the teacher are matched or how they are compared. A few logit-based methods also exploit relations among samples. DIST replaces the KL divergence with Pearson correlations computed both across the classes of each sample and across the samples of a batch for each class \citep{dist}, and MLKD aligns predictions at the instance, batch, and class levels, where the batch level matches the correlations among the predictions of different inputs \citep{mlkd}. HACD also transfers relations among samples. It measures them on centered, temperature-softened probability directions, which are invariant to additive logit offsets, and transfers them through instance discrimination, in which each student output must identify its own teacher target among the targets of the other samples, rather than by matching correlation statistics as DIST and MLKD do. A related line of work studies data augmentation for KD: \citet{patient} show that teacher and student should see consistent inputs, \citet{goodaugmentation} analyze from a statistical perspective which augmentations benefit distillation, and CRLD feeds weakly and strongly augmented views to both the teacher and the student, regularizes within-view and cross-view consistency, and filters teacher predictions by confidence \citep{crld}.
\textbf{Feature-based distillation.} Feature-based methods transfer intermediate representations, such as hint features \citep{finets}, attention maps \citep{attetion}, pairwise similarities \citep{similarity}, relational structures \citep{relational}, multi-stage features \citep{review}, and class attention maps \citep{catkd}. Their supervision relies on a meaningful correspondence between teacher and student features, which becomes difficult to establish when the two networks belong to different architectural families. The same limitation applies to methods that transfer relations among samples, such as SP \citep{similarity} and RKD \citep{relational}, because the relations are still computed from architecture-dependent features.

\subsection{Heterogeneous Knowledge Distillation}

Heterogeneous knowledge distillation transfers knowledge between CNNs, Transformers, and MLP-based models, whose internal representations can differ substantially. Most existing methods bridge this gap through feature adaptation. CAKD aligns heterogeneous features through partial cross-attention and group-wise linear projectors \citep{cakd}. OFA-KD (OFA) projects intermediate student features into the logit space through auxiliary exit branches and adaptively enhances the target information \citep{ofa}. PAT adapts teacher features to the student through prompt-tuning blocks driven by student feedback, together with region-aware attention that mitigates the view mismatch between architectures \citep{pat}. RSD suppresses architecture-exclusive information through cross-architecture invariance and feature decorrelation, using a lightweight module that decouples this objective from the representations of the student \citep{rsd}, and HCD maps fused teacher and student features to shared logits through a complementary feature mapper and decomposes them into diverse sub-logits \citep{hcd}. HACD uses only the native classifier logits, which share the same label space regardless of the architecture, and transfers the relations among them.

\subsection{Contrastive Learning and Contrastive Distillation}

Contrastive learning learns representations by distinguishing positive pairs from negative samples. Instance discrimination and contrastive predictive coding establish instance-level objectives \citep{instdisc,infonce}, SimCLR and MoCo~v2 show that strong augmentations and projection heads are central to contrastive learning, with MoCo~v2 maintaining a large set of negatives through a momentum encoder and a queue \citep{simclr,moco}, and CLIP aligns paired image and text embeddings with a bidirectional contrastive objective \citep{clip}. In knowledge distillation, CRD contrasts teacher and student features with projection heads and a memory bank \citep{crd}, and CKD treats the teacher and student logits of the same sample as a positive pair and the logits of the other samples in the batch as negatives, optimizing an InfoNCE objective on $\ell_2$-normalized logits \citep{ckd}.

CRD \citep{crd} argued that the KL objective of KD \citep{kd} treats the output dimensions independently and ignores the structural knowledge of the teacher representation, which it captures by maximizing a contrastive lower bound on the teacher--student mutual information, and CKD \citep{ckd} brought sample-wise contrast to logits, jointly enforcing intra-sample alignment and inter-sample separation. Both were developed and evaluated mainly for homogeneous teacher-student pairs. HACD uses contrast because, in the heterogeneous setting, a student from another architectural family should inherit the relations among teacher outputs, whose values are architecture-specific. This leads to three differences from CRD and CKD. First, the relations are measured on centered, temperature-softened probability directions with analytically characterized properties, rather than on features with projection heads and a memory bank, as in CRD, or on $\ell_2$-normalized raw logits, as in CKD, which change under an additive logit offset; the representation of HACD is offset-invariant and reduces to the centered logit direction as $T\to\infty$ (Proposition~\ref{prop:limits}). Second, HACD contrasts the student and teacher outputs in both directions, following CLIP \citep{clip}. Third, because every teacher target affects the supervision of the whole batch, HACD computes the targets on the moderate view only and aligns both student views with them, whereas CRLD \citep{crld} passes both views through the teacher, aligns the student within and across views, and selects reliable teacher predictions by confidence; HACD thus needs a single teacher pass per sample.

\begin{figure*}[t]
    \centering
    \includegraphics[clip, trim=5 5 5 3, width=\textwidth]{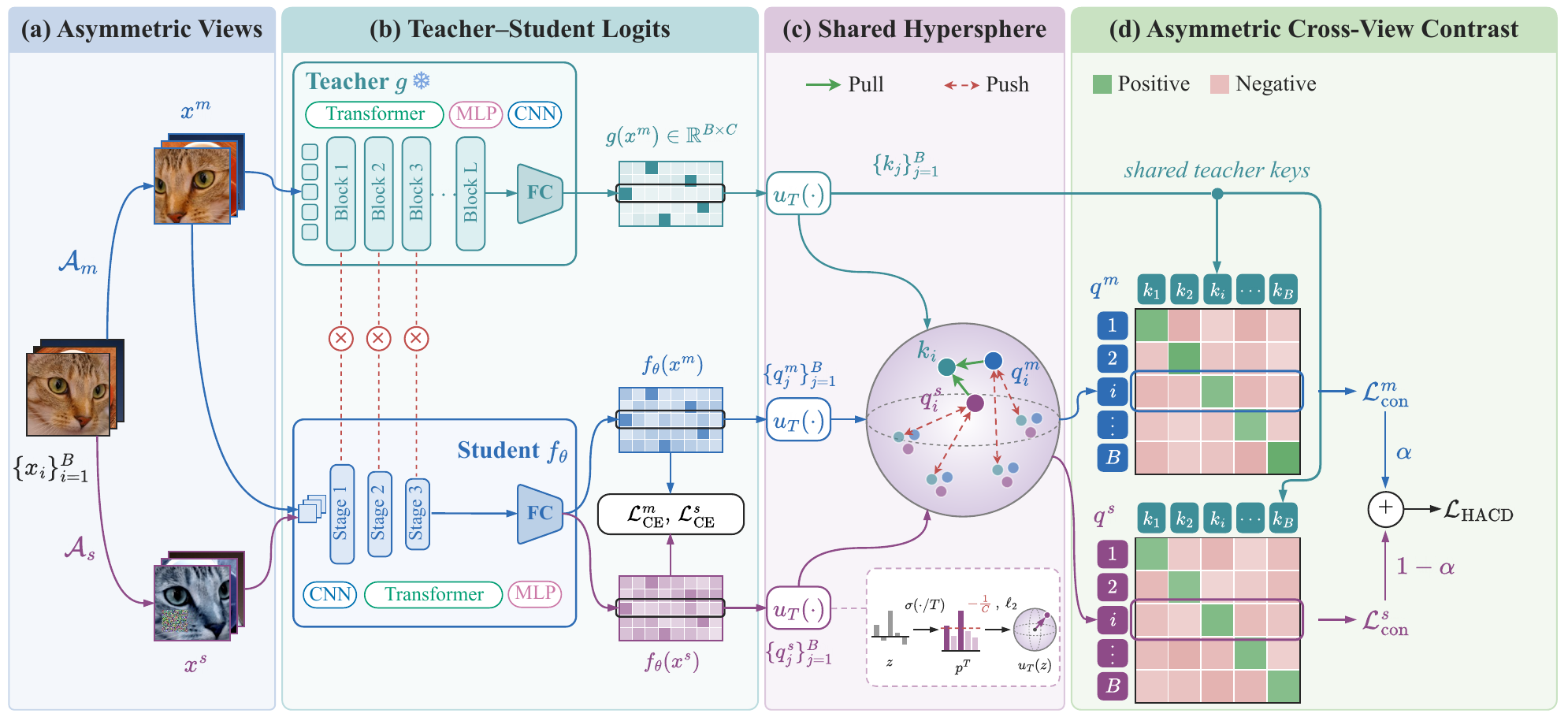}
    \caption{Overview of HACD. Each training image is augmented into a moderate view and a strong view. The frozen teacher processes only the moderate view, whereas the student processes both views. Teacher and student logits are mapped to centered, temperature-softened probability directions on a shared unit hypersphere (Eq.~\eqref{eq:logit_normalization}). The same teacher representations are the targets of bidirectional instance-level contrast for both student views (Eq.~\eqref{eq:bidirectional_contrastive}), and the two view losses are combined with weight $\alpha$ (Eq.~\eqref{eq:overall_loss}). Only the student is used at inference.}
    \label{fig:overview}
\end{figure*}

\section{Method}
\label{sec:method}

\subsection{Preliminaries and Overview}
\label{sec:preliminaries}

Let $g$ denote a frozen teacher network and $f_\theta$ a student network with parameters $\theta$. The two networks may belong to different architectural families, e.g., a Transformer teacher and a CNN student, but share the same $C$-class output space. Given a mini-batch of labeled samples $\{(x_i,y_i)\}_{i=1}^{B}$, HACD transfers knowledge exclusively through the logits of the two networks, without accessing intermediate features or introducing feature alignment modules.

Fig.~\ref{fig:overview} illustrates the framework, whose three components implement the design choices of Section~\ref{sec:intro}. The \emph{asymmetric view construction} (Section~\ref{sec:asymmetric_views}) keeps the teacher targets reliable: each image yields a moderately and a strongly augmented view, the teacher processes only the former, and the student processes both. The \emph{temperature-softened probability representation} (Section~\ref{sec:logit_representation}) makes teacher and student relations comparable by mapping both sets of logits onto a shared unit hypersphere. The \emph{asymmetric cross-view contrastive distillation} (Section~\ref{sec:asymmetric_contrastive}) transfers the relations through bidirectional instance-level contrast between the same teacher representations and both student views. We describe the view construction first because it defines the inputs of the other two components; Section~\ref{sec:training_objective} gives the overall objective and the training procedure.

\subsection{Asymmetric View Construction}
\label{sec:asymmetric_views}

Strong data augmentation increases view diversity and is central to contrastive learning \citep{simclr,cmc}. In knowledge distillation, however, teacher predictions on strongly augmented inputs can be less reliable, which CRLD addresses by filtering them by confidence \citep{crld}, and the benefit of augmentation depends on how the teacher and student inputs correspond \citep{patient,goodaugmentation}. This issue matters more when the relations among teacher outputs are the supervision: in a contrastive objective, each teacher target is the positive of one sample and a negative of all others, so unreliable targets, when they are the primary supervision, can distort the relations of the entire batch. HACD therefore separates the source of input diversity from the source of supervision through an asymmetric view construction. Given an image $x_i$, two views are generated as
\begin{equation}
    x_i^{m} = a_m(x_i),\quad a_m \sim \mathcal{A}_m,
    \quad
    x_i^{s} = a_s(x_i),\quad a_s \sim \mathcal{A}_s,
\label{eq:asymmetric_view_generation}
\end{equation}
where $\mathcal{A}_m$ denotes a moderate augmentation distribution consisting of geometric and photometric transformations that preserve discriminative content, and $\mathcal{A}_s$ denotes a strong augmentation distribution that shares the geometric transformations of $\mathcal{A}_m$ but replaces its photometric transformations with stronger, randomly sampled photometric and mild geometric perturbations. The teacher processes only $x_i^{m}$ and runs a single forward pass per image. The student processes both $x_i^{m}$ and $x_i^{s}$, and its two forward and backward passes are the main training overhead of HACD (Section~\ref{sec:ablation}). The teacher targets thus come from moderate perturbations, and strong perturbations only diversify the inputs of the student. Both student views are aligned with the same targets (Section~\ref{sec:asymmetric_contrastive}). The augmentation configurations are detailed in Section~\ref{sec:experimental_setup}.

\subsection{Temperature-Softened Probability Representation}
\label{sec:logit_representation}

Relations between teacher and student outputs can be compared only if both are expressed in a common space. The logits of heterogeneous architectures, however, can differ substantially in their additive offsets and overall magnitudes \citep{lskd}, as well as in their sharpness (Fig.~\ref{fig:logit_statistics}). A common representation for both should be invariant to additive logit offsets, place the teacher and the student on a common norm, and retain the relative structure among non-target classes, i.e., the dark knowledge of the teacher \citep{kd}. To this end, given a logit vector $z\in\mathbb{R}^{C}$ and a class temperature $T>0$, we define:
\begin{equation}
    p^{T}(z)=\operatorname{softmax}\!\left(z/T\right),\quad
    u_T(z)=\frac{\Pi\,p^{T}(z)}{\big\|\Pi\,p^{T}(z)\big\|_2},\quad
    \Pi = I-\tfrac{1}{C}\mathbf{1}\mathbf{1}^{\top},
\label{eq:logit_normalization}
\end{equation}
where $\mathbf{1}\in\mathbb{R}^{C}$ denotes the all-ones vector and $\Pi$ is the projection onto the orthogonal complement of $\mathbf{1}$. Since $\mathbf{1}^{\top}p^{T}(z)=1$, we have $\Pi\,p^{T}(z)=p^{T}(z)-\tfrac{1}{C}\mathbf{1}$; that is, centering subtracts the uniform distribution. The teacher and the student share this mapping:
\begin{equation}
    k_i = u_T\!\left(g(x_i^{m})\right),\quad
    q_i^{v} = u_T\!\left(f_\theta(x_i^{v})\right),\quad v \in \{m,s\},
\label{eq:teacher_student_directions}
\end{equation}
where no gradient is propagated through the teacher representation $k_i$. The class temperature $T$ determines how the softened probability mass is distributed over the classes and, consequently, the geometry of the representation. \ref{app:theory} characterizes this dependence formally. For every $T$, the mapping $u_T$ is invariant to additive logit offsets and has unit norm. It is not invariant to a rescaling of the logits for finite $T$, because rescaling changes the sharpness of $p^{T}$; it becomes scale-invariant only in the limit $T\to\infty$, where it reduces to the centered logit direction (Proposition~\ref{prop:limits}). The temperature therefore controls how much of the sharpness of a teacher enters the representation. The centering $\Pi$ is indispensable: without it, the representations of different samples collapse towards a common direction whenever the softened predictions are close to uniform (Proposition~\ref{prop:centering}). As $T$ varies, $u_T$ moves between two limits, namely the centered logit direction as $T\to\infty$ and the one-hot direction of the predicted class as $T\to0$ (Proposition~\ref{prop:limits}), while the effective number of classes, $N_{\mathrm{eff}}=\big(\sum_{c}(p^{T}_{c})^{2}\big)^{-1}$, increases monotonically with $T$ and the sensitivity of the representation to class $k$ scales with $p^{T}_{k}$ (Proposition~\ref{prop:support}).

Neither limit suits instance-level contrast. In the high-temperature limit, every class contributes in proportion to its centered logit, so a teacher that spreads its output over a long tail of non-target classes yields a representation dominated by this tail, whose ordering carries little semantic information. In the low-temperature limit, samples with the same predicted class collapse to nearly the same representation, which removes the intra-class variation that distinguishes individual samples (Proposition~\ref{prop:limits}(b)). A moderate temperature avoids both cases. Since the sensitivity of $u_T$ to class $k$ scales with $p^{T}_{k}$ (Proposition~\ref{prop:support}), the tail classes, which receive little probability, barely change the representation, whereas the top-ranked non-target classes, i.e., the dark knowledge, still shape its direction and keep samples of the same class apart. Heterogeneous teachers make this choice harder, because their sharpness differs across architectures (Fig.~\ref{fig:logit_statistics}): the same temperature may leave a long tail for one teacher and a nearly one-hot output for another. Because $p^{T}$ coincides with the softened target of KD \citep{kd}, $T$ plays the same role here as in KD \citep{ctkd}: we adopt $T=4$, the value widely used in logit distillation \citep{crd,dkd,dist}, when the teacher output has a long tail, and the centered logit direction ($T\to\infty$) when it already concentrates on a few classes. Section~\ref{sec:ablation} analyzes this choice on ImageNet-1K teachers.

\subsection{Asymmetric Cross-View Contrastive Distillation}
\label{sec:asymmetric_contrastive}
\textbf{Bidirectional instance contrast.} Per-sample distribution matching asks the student to reproduce each teacher output, including its architecture-specific sharpness, and does not model the relationships among samples. HACD instead asks the student to arrange its outputs as the teacher arranges its own. Following the bidirectional formulation of CLIP \citep{clip}, the student must identify its corresponding teacher representation within the mini-batch. For student view $v\in\{m,s\}$, the similarity matrix and the contrastive loss are defined as:
\begin{equation}
    S_{ij}^{v}=\frac{(q_i^{v})^{\top}k_j}{\tau},\quad i,j\in\{1,\ldots,B\},
\label{eq:contrastive_similarity}
\end{equation}
\begin{equation}
    \mathcal{L}_{\mathrm{con}}^{v}
    =-\frac{1}{2B}\sum_{i=1}^{B}
    \left[
    \log\frac{\exp(S_{ii}^{v})}{\sum_{j=1}^{B}\exp(S_{ij}^{v})}
    +\log\frac{\exp(S_{ii}^{v})}{\sum_{j=1}^{B}\exp(S_{ji}^{v})}
    \right],
\label{eq:bidirectional_contrastive}
\end{equation}
where $\tau$ is the instance temperature, which controls the emphasis on hard negatives \citep{contrastiveloss}; following MoCo~v2 \citep{moco}, we set $\tau=0.2$. The two terms in Eq.~\eqref{eq:bidirectional_contrastive} perform student-to-teacher and teacher-to-student matching, respectively. Since $\|q_i^{v}\|_2=\|k_i\|_2=1$, the positive similarity can be written as $S_{ii}^{v}=\big(1-\tfrac{1}{2}\|q_i^{v}-k_i\|_2^{2}\big)/\tau$. The positive term therefore still aligns each student output with its own teacher target, but in direction on the shared hypersphere rather than in the probability values of the softened distribution. The negative terms require each student output to be closer to its own target than to the targets of the other samples, which transfers the inter-sample relations of the teacher; per-sample distillation has no such term. Because the loss depends on the similarities of a student output relative to all teacher targets in the batch, it does not require the student to reproduce the probability values of a heterogeneous teacher, although, for finite $T$, the sharpness of the teacher still shapes the direction of its targets. Samples of the same class are not excluded from the negatives. Because such samples share the target-class component, they can be distinguished only through their non-target structure, which compels the student to inherit the fine-grained intra-class relations of the teacher; the tolerance to such semantically similar negatives is governed by $\tau$ \citep{contrastiveloss}.

\textbf{Asymmetric cross-view supervision.} The same teacher representation $k_i$ is the contrastive target of both $q_i^{m}$ and $q_i^{s}$. Applying Eq.~\eqref{eq:bidirectional_contrastive} with $v=m$ and $v=s$ yields $\mathcal{L}_{\mathrm{con}}^{m}$ and $\mathcal{L}_{\mathrm{con}}^{s}$, respectively. The former transfers instance relations under controlled perturbations. The latter requires the student to recover the same relations under strong perturbations, while its targets still originate from the moderate view. Teacher predictions on strongly augmented inputs are less reliable, and if they served as the targets of both views, they would affect every sample through the negatives (Table~\ref{tab:ablation_components}, row (e)); computing the targets on the moderate view keeps them out of the primary supervision. The moderate view aligns the student with $k_i$ under perturbations that the teacher also sees, and the strong view extends this alignment to perturbations that only the student sees. Because the targets of the strong view come from the teacher and not from the predictions of the student, the student is not driven towards targets that deviate from the teacher.

\begin{algorithm}[t]
\caption{Training procedure of HACD}
\label{alg:hacd}
\begin{algorithmic}[1]
\Require Frozen teacher $g$; student $f_\theta$; augmentation distributions $\mathcal{A}_m,\mathcal{A}_s$; hyperparameters $T,\tau,\alpha,\lambda,E_{\mathrm{ce}}$
\For{each epoch $e$}
    \For{each mini-batch $\{(x_i,y_i)\}_{i=1}^{B}$}
        \State Generate $x_i^{m}$ and $x_i^{s}$ by Eq.~\eqref{eq:asymmetric_view_generation}
        \State Compute $k_i=u_T\big(g(x_i^{m})\big)$ without gradient \Comment{Eq.~\eqref{eq:teacher_student_directions}}
        \For{$v\in\{m,s\}$}
            \State Compute $q_i^{v}=u_T\big(f_\theta(x_i^{v})\big)$ and $\mathcal{L}^{v}$ \Comment{Eqs.~\eqref{eq:contrastive_similarity}--\eqref{eq:view_loss}}
        \EndFor
        \State Compute $\mathcal{L}_{\mathrm{HACD}}$ by Eq.~\eqref{eq:overall_loss} and update $\theta\leftarrow\theta-\eta\nabla_{\theta}\mathcal{L}_{\mathrm{HACD}}$
    \EndFor
\EndFor
\Ensure Trained student $f_\theta$
\end{algorithmic}
\end{algorithm}

\subsection{Training Objective}
\label{sec:training_objective}

For each view $v\in\{m,s\}$, the student loss consists of a contrastive term and a label term:
\begin{equation}
    \mathcal{L}^{v}=\mathcal{L}_{\mathrm{con}}^{v}+\lambda\,w(e)\,\mathcal{L}_{\mathrm{CE}}^{v},
    \quad
    \mathcal{L}_{\mathrm{CE}}^{v}=-\frac{1}{B}\sum_{i=1}^{B}\log\sigma\big(f_\theta(x_i^{v})\big)_{y_i},
\label{eq:view_loss}
\end{equation}
where $\sigma(\cdot)$ denotes the softmax function and $\lambda\in[0,1]$ controls the strength of label supervision. The weight of the contrastive term is fixed to $1$ and sets the reference scale. The label weight increases linearly with the training epoch $e$,
\begin{equation}
    w(e)=\min\!\left(e/E_{\mathrm{ce}},\,1\right),
\label{eq:ce_weight}
\end{equation}
so that the relations transferred from the teacher dominate the early stage of training. The overall objective is a convex combination of the two views:
\begin{equation}
    \mathcal{L}_{\mathrm{HACD}}=\alpha\,\mathcal{L}^{m}+(1-\alpha)\,\mathcal{L}^{s},\quad \alpha\in[0,1].
\label{eq:overall_loss}
\end{equation}

Apart from $T$ and $\tau$, which follow prior work, HACD introduces only three hyperparameters: the view weight $\alpha$, the label strength $\lambda$, and the warm-up length $E_{\mathrm{ce}}$. Setting $\alpha=1$ reduces HACD to single-view contrast, and setting $\lambda=0$ yields label-free distillation. Their values are given in Section~\ref{sec:experimental_setup}, and their sensitivity and the training cost are analyzed in Section~\ref{sec:ablation}. Algorithm~\ref{alg:hacd} summarizes the training procedure. At inference, only the student is used, so HACD introduces no additional parameters or inference cost.

\begin{table*}[t]
\centering
\caption{
Top-1 classification accuracy (\%) of cross-architecture knowledge
distillation on CIFAR-100.
All 12 teacher-student pairs involve heterogeneous architectures.
The best and second-best results are highlighted in
\textbf{bold} and \underline{underline}, respectively. T. and S. denote the accuracies of the teacher and student models before distillation, respectively. Average Gain represents the mean accuracy improvement over the corresponding student trained from scratch, calculated across all 12 teacher-student pairs.
$\Delta$ represents the improvement of HACD over vanilla KD.
Average Gain and $\Delta$ are absolute differences in percentage points (pp).
$^{*}$ indicates results reproduced by LDRLD~\citep{ldrld}.
``--'' indicates results not reported for the corresponding
teacher-student pair.
HCD~\citep{hcd} is excluded from the Average Gain comparison because
results are available for only six pairs.}
\label{tab:cifar100_results}

\resizebox{\textwidth}{!}{%
\begin{tabular}{@{}llcc*{6}{c}*{8}{c}c@{}}
\toprule

\multicolumn{2}{c}{Architecture}
&
\multicolumn{2}{c}{From Scratch}
&
\multicolumn{6}{c}{Feature-based}
&
\multicolumn{8}{c}{Logit-based}
&
\multicolumn{1}{c}{Gain}
\\

\cmidrule(lr){1-2}
\cmidrule(lr){3-4}
\cmidrule(lr){5-10}
\cmidrule(lr){11-18}
\cmidrule(l){19-19}

Teacher
&
Student
&
T.
&
S.
&
\shortstack{FitNet\\\citep{finets}}
&
\shortstack{CC\\\citep{correlation}}
&
\shortstack{RKD\\\citep{relational}}
&
\shortstack{CRD\\\citep{crd}}
&
\shortstack{PAT\\\citep{pat}}
&
\shortstack{RSD\\\citep{rsd}}
&
\shortstack{KD\\\citep{kd}}
&
\shortstack{DKD\\\citep{dkd}}
&
\shortstack{DIST\\\citep{dist}}
&
\shortstack{OFA\\\citep{ofa}}
&
\shortstack{TeKAP\\\citep{tekap}}
&
\shortstack{LDRLD\\\citep{ldrld}}
&
\shortstack{HCD\\\citep{hcd}}
&
\textbf{HACD}
&
$\Delta$
\\

\midrule

\rowcolor{gray!15}
\multicolumn{19}{c}{\textit{CNN-based students}}
\\

Swin-T
& ResNet-18
& 89.26 & 74.01
& 78.87 & 74.19 & 74.11 & 77.63 & 81.22 & \underline{83.92}
& 78.74 & 80.26 & 77.75 & 80.54
& 81.38$^{*}$ & 82.17 & 82.78
& \cellcolor{gray!20}\textbf{85.00}
& \textcolor{green!50!black}{+6.26}
\\

ViT-S
& ResNet-18
& 92.04 & 74.01
& 77.71 & 74.26 & 73.72 & 76.60 & 80.11 & \underline{81.50}
& 77.26 & 78.10 & 76.49 & 80.15
& 79.06$^{*}$ & 80.36 & 81.33
& \cellcolor{gray!20}\textbf{84.76}
& \textcolor{green!50!black}{+7.50}
\\

Mixer-B/16
& ResNet-18
& 87.29 & 74.01
& 77.15 & 74.26 & 73.75 & 76.42 & 80.07 & \underline{81.85}
& 77.79 & 78.67 & 76.36 & 79.39
& 80.05$^{*}$ & 80.69 & 81.53
& \cellcolor{gray!20}\textbf{83.95}
& \textcolor{green!50!black}{+6.16}
\\

Swin-T
& MobileNetV2
& 89.26 & 73.68
& 74.28 & 71.19 & 69.00 & 79.80 & 78.78 & \underline{83.68}
& 74.68 & 71.07 & 72.89 & 80.98
& 80.23$^{*}$ & 81.64 & 82.19
& \cellcolor{gray!20}\textbf{85.38}
& \textcolor{green!50!black}{+10.70}
\\

ViT-S
& MobileNetV2
& 92.04 & 73.68
& 73.54 & 70.67 & 68.46 & 78.14 & 78.87 & \underline{81.68}
& 72.77 & 69.80 & 72.54 & 78.45
& 78.41$^{*}$ & 79.21 & 80.81
& \cellcolor{gray!20}\textbf{84.66}
& \textcolor{green!50!black}{+11.89}
\\

Mixer-B/16
& MobileNetV2
& 87.29 & 73.68
& 73.78 & 70.73 & 68.95 & 78.15 & 78.62 & \underline{81.74}
& 73.33 & 70.20 & 73.26 & 78.78
& 79.89$^{*}$ & 80.64 & 81.09
& \cellcolor{gray!20}\textbf{84.20}
& \textcolor{green!50!black}{+10.87}
\\

\midrule

\rowcolor{gray!15}
\multicolumn{19}{c}{\textit{ViT-based students}}
\\

ConvNeXt-T
& DeiT-T
& 88.41 & 68.00
& 60.78 & 68.01 & 69.79 & 65.94 & 79.59 & \underline{82.46}
& 72.99 & 74.60 & 73.55 & 75.76
& 76.32$^{*}$ & 77.46 & --
& \cellcolor{gray!20}\textbf{82.50}
& \textcolor{green!50!black}{+9.51}
\\

Mixer-B/16
& DeiT-T
& 87.29 & 68.00
& 71.05 & 68.13 & 69.89 & 65.35 & 74.66 & \underline{78.50}
& 71.36 & 73.44 & 71.67 & 73.90
& 74.83$^{*}$ & 75.31 & --
& \cellcolor{gray!20}\textbf{81.46}
& \textcolor{green!50!black}{+10.10}
\\

ConvNeXt-T
& Swin-P
& 88.41 & 72.63
& 24.06 & 72.63 & 71.73 & 67.09 & 80.74 & \underline{82.21}
& 76.44 & 76.80 & 76.41 & 78.32
& 79.18$^{*}$ & 80.71 & --
& \cellcolor{gray!20}\textbf{82.46}
& \textcolor{green!50!black}{+6.02}
\\

Mixer-B/16
& Swin-P
& 87.29 & 72.63
& 75.20 & 73.32 & 70.82 & 67.03 & 78.44 & \underline{81.28}
& 75.93 & 76.39 & 75.85 & 78.93
& 78.97$^{*}$ & 80.52 & --
& \cellcolor{gray!20}\textbf{82.84}
& \textcolor{green!50!black}{+6.91}
\\

\midrule

\rowcolor{gray!15}
\multicolumn{19}{c}{\textit{MLP-based students}}
\\

ConvNeXt-T
& ResMLP-S12
& 88.41 & 66.56
& 45.47 & 67.70 & 65.82 & 63.35 & \underline{83.50} & \textbf{84.21}
& 72.25 & 73.22 & 71.93 & 81.22
& 81.14$^{*}$ & 79.28 & --
& \cellcolor{gray!20}81.84
& \textcolor{green!50!black}{+9.59}
\\

Swin-T
& ResMLP-S12
& 89.26 & 66.56
& 63.12 & 68.37 & 64.66 & 61.72 & 80.94 & \textbf{82.67}
& 71.89 & 72.82 & 11.05 & 80.63
& 80.22$^{*}$ & 80.54 & --
& \cellcolor{gray!20}\underline{82.23}
& \textcolor{green!50!black}{+10.34}
\\

\midrule

\multicolumn{4}{l}{\textbf{Average Gain}}
&
\textcolor{red!70!black}{-5.20}
&
\textcolor{red!70!black}{-0.33}
&
\textcolor{red!70!black}{-1.40}
&
\textcolor{red!70!black}{-0.02}
&
\textcolor{green!50!black}{+8.17}
&
\textcolor{green!50!black}{\underline{+10.69}}
&
\textcolor{green!50!black}{+3.17}
&
\textcolor{green!50!black}{+3.16}
&
\textcolor{red!70!black}{-2.31}
&
\textcolor{green!50!black}{+7.47}
&
\textcolor{green!50!black}{+7.69}
&
\textcolor{green!50!black}{+8.42}
&
--
&
\cellcolor{gray!20}\textcolor{green!80!black}{\textbf{+11.99}}
&
\\

\bottomrule
\end{tabular}
}
\end{table*}

\section{Experiments}
\label{sec:experiments}

\subsection{Experimental Setup}
\label{sec:experimental_setup}
Following the experimental protocol of OFA \citep{ofa}, we evaluate HACD on heterogeneous teacher-student pairs spanning CNN, Transformer, and MLP architectures. We adopt its model configurations and architecture-specific training strategies, with the adaptations described below.

\textbf{Datasets.} CIFAR-100 \citep{cifar} contains 50,000 training and 10,000 test images of 100 classes at $32\times32$ pixels; following OFA, we resize them to $224\times224$ to accommodate all architectures. ImageNet-1K \citep{imagenet} contains about 1.28 million training and 50,000 validation images of 1,000 classes, processed at $224\times224$.

\textbf{Models.} Following OFA~\citep{ofa}, we evaluate knowledge transfer across architectural families, covering CNNs (ResNet~\citep{resnet}, MobileNetV2~\citep{mobilenetv2}, and ConvNeXt~\citep{convnext}), Transformers (ViT~\citep{vit}, DeiT~\citep{deit}, and Swin Transformer~\citep{swin}), and MLP-based models (MLP-Mixer~\citep{mlpmixer} and ResMLP~\citep{resmlp}).

On CIFAR-100, we evaluate 12 heterogeneous teacher-student pairs, with Swin-T, ViT-S, Mixer-B/16, and ConvNeXt-T as teachers and ResNet-18, MobileNetV2, DeiT-T, Swin-P, and ResMLP-S12 as students. On ImageNet-1K, we evaluate 15 pairs: DeiT-T, Swin-T, and Mixer-B/16 as teachers for ResNet-18 and MobileNetV2; ResNet-50, ConvNeXt-T, and Mixer-B/16 as teachers for DeiT-T and Swin-N; and ResNet-50, ConvNeXt-T, and Swin-T as teachers for ResMLP-S12. Swin-P and Swin-N denote the reduced Swin variants used by OFA for heterogeneous distillation on CIFAR-100 and ImageNet-1K, respectively~\citep{ofa}.

\textbf{Baselines.} Logit-based methods include vanilla KD \citep{kd}, DKD \citep{dkd}, DIST \citep{dist}, and TeKAP \citep{tekap}. Feature-based approaches include FitNets \citep{finets}, CC \citep{correlation}, RKD \citep{relational}, and CRD \citep{crd}. Heterogeneous methods include OFA \citep{ofa}, PAT \citep{pat}, RSD \citep{rsd}, HCD \citep{hcd}, and LDRLD \citep{ldrld}. Baseline results are taken mainly from the published heterogeneous evaluations of LDRLD, HCD, and RSD \citep{ldrld,hcd,rsd} under their own training configurations; KD under the configuration of HACD is reported in Section~\ref{sec:ablation}. We report top-1 accuracy.

\textbf{Implementation Details.} Following OFA, CNN and Transformer/MLP students use separate optimization settings. Teachers are pretrained on each target dataset and kept frozen throughout distillation, while students are trained from scratch. For students with zero-initialized classification heads, such as ResMLP, we reinitialize the classifier using a truncated normal distribution ($\sigma=0.02$) to avoid degenerate logit normalization at initialization. On both datasets, Transformer/MLP students use AdamW (learning rate $5\times10^{-4}$, weight decay $0.05$), gradient clipping of $5.0$, Mixup of $0.8$, CutMix of $1.0$, and random erasing with probability $0.25$.

\textit{On CIFAR-100}, all students are trained for 300 epochs on four GPUs with 128 images per GPU. CNN students use SGD with momentum $0.9$, learning rate $0.04$, and weight decay $2\times10^{-3}$, together with cosine scheduling and random erasing with probability $0.1$. When Mixup or CutMix is applied, the mixing configuration is shared across the corresponding teacher and student views to preserve input correspondence.

\textit{On ImageNet-1K}, all experiments use eight GPUs with 128 images per GPU. CNN students are trained for 100 epochs using SGD with momentum $0.9$, learning rate $0.3$, weight decay $10^{-4}$, and cosine scheduling. Transformer/MLP students are trained for 300 epochs, with an additional 20 warm-up epochs and label smoothing of $0.1$ beyond the shared settings above. No model EMA is used. Training uses mixed precision, with bfloat16 for Transformer/MLP students to avoid overflow in ResMLP, and evaluation follows the standard timm pipeline.

\begin{table*}[t]
\centering
\caption{
Top-1 classification accuracy (\%) of cross-architecture knowledge
distillation on ImageNet-1K. Notation follows Table~\ref{tab:cifar100_results}.
Average Gain is computed over the 15 pairs and reported only for methods with complete results.
$^{*}$ indicates results obtained by combining the corresponding method
with FitNets, following OFA.
$^{\dagger}$ indicates results taken from HCD~\citep{hcd}.
}
\label{tab:imagenet_results}

\resizebox{\textwidth}{!}{%
\begin{tabular}{@{}llcc*{6}{c}*{7}{c}c@{}}
\toprule

\multicolumn{2}{c}{Architecture}
&
\multicolumn{2}{c}{From Scratch}
&
\multicolumn{6}{c}{Feature-based}
&
\multicolumn{7}{c}{Logit-based}
&
\multicolumn{1}{c}{Gain}
\\

\cmidrule(lr){1-2}
\cmidrule(lr){3-4}
\cmidrule(lr){5-10}
\cmidrule(lr){11-17}
\cmidrule(l){18-18}

Teacher
&
Student
&
T.
&
S.
&
\shortstack{FitNet\\\citep{finets}}
&
\shortstack{CC\\\citep{correlation}}
&
\shortstack{RKD\\\citep{relational}}
&
\shortstack{CRD\\\citep{crd}}
&
\shortstack{PAT\\\citep{pat}}
&
\shortstack{RSD\\\citep{rsd}}
&
\shortstack{KD\\\citep{kd}}
&
\shortstack{DKD\\\citep{dkd}}
&
\shortstack{DIST\\\citep{dist}}
&
\shortstack{OFA\\\citep{ofa}}
&
\shortstack{TeKAP\\\citep{tekap}}
&
\shortstack{HCD\\\citep{hcd}}
&
\textbf{HACD}
&
$\Delta$
\\

\midrule

\rowcolor{gray!15}
\multicolumn{18}{c}{\textit{CNN-based students}}
\\

DeiT-T
& ResNet-18
& 72.19 & 69.75
& 70.44 & 69.77 & 69.47 & 69.25 & -- & \textbf{71.70}
& 70.22 & 69.39 & 70.64 & 71.34
& -- & --
& \underline{71.50} \cellcolor{gray!20}
& \textcolor{green!50!black}{+1.28}
\\

Swin-T
& ResNet-18
& 81.38 & 69.75
& 71.18 & 70.07 & 68.89 & 69.09 & 71.54 & \underline{72.13}
& 71.14 & 71.10 & 70.91 & 71.85
& 71.25$^{\dagger}$ & 71.91
& \textbf{72.76} \cellcolor{gray!20}
& \textcolor{green!50!black}{+1.62}
\\

Mixer-B/16
& ResNet-18
& 76.62 & 69.75
& 70.78 & 70.05 & 69.46 & 68.40 & -- & 71.41
& 70.89 & 69.89 & 70.66 & 71.38
& 70.95$^{\dagger}$ & \underline{71.66}
& \textbf{71.81} \cellcolor{gray!20}
& \textcolor{green!50!black}{+0.92}
\\

DeiT-T
& MobileNetV2
& 72.17 & 68.87
& 70.95 & 70.69 & 69.72 & 69.60 & -- & \underline{72.18}
& 70.87 & 70.14 & 71.08 & 71.39
& 70.23$^{\dagger}$ & 71.38
& \textbf{72.88} \cellcolor{gray!20}
& \textcolor{green!50!black}{+2.01}
\\

Swin-T
& MobileNetV2
& 81.38 & 68.87
& 71.75 & 70.69 & 67.52 & 69.58 & -- & 72.36
& 72.05 & 71.71 & 71.76 & 72.32
& 72.50$^{\dagger}$ & \underline{72.72}
& \textbf{73.38} \cellcolor{gray!20}
& \textcolor{green!50!black}{+1.33}
\\

Mixer-B/16
& MobileNetV2
& 76.62 & 68.87
& 71.59 & 70.79 & 69.86 & 68.89 & 72.22 & 71.90
& 71.92 & 70.93 & 71.74 & 72.12
& 72.01$^{\dagger}$ & \underline{72.32}
& \textbf{73.13} \cellcolor{gray!20}
& \textcolor{green!50!black}{+1.21}
\\

\midrule

\rowcolor{gray!15}
\multicolumn{18}{c}{\textit{ViT-based students}}
\\

ResNet-50
& DeiT-T
& 80.33 & 72.17
& 75.84 & 72.56 & 72.06 & 68.53 & -- & \underline{76.17}$^{*}$
& 75.10 & 75.60$^{*}$ & 75.13$^{*}$ & \textbf{76.55}$^{*}$
& -- & --
& \underline{76.17} \cellcolor{gray!20}
& \textcolor{green!50!black}{+1.07}
\\

ConvNeXt-T
& DeiT-T
& 82.05 & 72.17
& 70.45 & 73.12 & 71.47 & 69.18 & 74.44 & \underline{74.46}
& 74.00 & 73.95 & 74.07 & 74.41
& -- & --
& \textbf{75.89} \cellcolor{gray!20}
& \textcolor{green!50!black}{+1.89}
\\

Mixer-B/16
& DeiT-T
& 76.55 & 72.17
& 74.38 & 72.82 & 72.24 & 68.23 & -- & 74.26
& 74.16 & 72.82 & 74.22 & \underline{74.46}
& -- & --
& \textbf{75.76} \cellcolor{gray!20}
& \textcolor{green!50!black}{+1.60}
\\

ResNet-50
& Swin-N
& 80.33 & 75.53
& 78.33 & 76.05 & 75.90 & 73.90 & -- & \textbf{78.78}$^{*}$
& 77.58 & 78.23$^{*}$ & 77.95$^{*}$ & \underline{78.64}$^{*}$
& -- & --
& 78.30 \cellcolor{gray!20}
& \textcolor{green!50!black}{+0.72}
\\

ConvNeXt-T
& Swin-N
& 82.05 & 75.53
& 74.81 & 75.79 & 75.48 & 74.15 & -- & \underline{77.70}
& 77.15 & 77.00 & 77.25 & 77.50
& -- & --
& \textbf{77.97} \cellcolor{gray!20}
& \textcolor{green!50!black}{+0.82}
\\

Mixer-B/16
& Swin-N
& 76.55 & 75.53
& 76.17 & 75.81 & 75.52 & 73.38 & -- & 76.55
& 76.26 & 75.03 & 76.54 & \underline{76.63}
& -- & --
& \textbf{77.00} \cellcolor{gray!20}
& \textcolor{green!50!black}{+0.74}
\\

\midrule

\rowcolor{gray!15}
\multicolumn{18}{c}{\textit{MLP-based students}}
\\

ResNet-50
& ResMLP-S12
& 80.33 & 76.65
& 78.13 & 76.21 & 75.45 & 73.23 & -- & 78.32$^{*}$
& 77.41 & 78.23$^{*}$ & 77.71$^{*}$ & \underline{78.53}$^{*}$
& -- & --
& \textbf{78.71} \cellcolor{gray!20}
& \textcolor{green!50!black}{+1.30}
\\

ConvNeXt-T
& ResMLP-S12
& 82.05 & 76.65
& 74.69 & 75.79 & 75.28 & 73.57 & -- & \underline{78.41}
& 76.84 & 77.23 & 77.24 & 77.53
& -- & --
& \textbf{78.65} \cellcolor{gray!20}
& \textcolor{green!50!black}{+1.81}
\\

Swin-T
& ResMLP-S12
& 81.35 & 76.65
& 76.48 & 76.15 & 75.10 & 73.40 & 77.59 & \underline{77.61}
& 76.67 & 76.99 & 77.25 & 77.31
& -- & --
& \textbf{78.35} \cellcolor{gray!20}
& \textcolor{green!50!black}{+1.68}
\\

\midrule

\multicolumn{4}{l}{\textbf{Average Gain}}
&
\textcolor{green!50!black}{+1.14}
&
\textcolor{green!50!black}{+0.50}
&
\textcolor{red!70!black}{-0.37}
&
\textcolor{red!70!black}{-1.77}
&
--
&
\textcolor{green!50!black}{\underline{+2.34}}
&
\textcolor{green!50!black}{+1.56}
&
\textcolor{green!50!black}{+1.29}
&
\textcolor{green!50!black}{+1.68}
&
\textcolor{green!50!black}{+2.20}
&
--
&
--
&
\cellcolor{gray!20}\textcolor{green!80!black}{\textbf{+2.89}}
&
\\

\bottomrule
\end{tabular}%
}
\end{table*}
\textbf{Augmentation \& Hyperparameters.} We use two views for contrastive distillation. The moderate view follows the augmentation pipeline of MoCo~v2~\citep{moco}. The strong view shares its geometric transformations and replaces its photometric transformations with stronger photometric and mild geometric perturbations randomly sampled for each image by RandAugment~\citep{randaugment}, following CRLD~\citep{crld}. HACD uses $\tau=0.2$ and $\alpha=0.7$ in all experiments. The class temperature and the label strength are set per dataset: $T=4$ and $\lambda=1$ on ImageNet-1K, as in the standard protocol of the compared methods \citep{kd,ofa}, with the label weight warmed up linearly over $E_{\mathrm{ce}}=30$ epochs, and the centered logit direction ($T\to\infty$, Proposition~\ref{prop:limits}) and $\lambda=0$, i.e., no cross-entropy term, on CIFAR-100. The choice of $T$ is analyzed in Section~\ref{sec:ablation}.

\subsection{Heterogeneous Distillation on CIFAR-100}
\label{sec:cifar100_results}

Table~\ref{tab:cifar100_results} reports the results on the 12 heterogeneous teacher-student pairs. HACD achieves the best accuracy on 10 pairs and the second-best on one. It attains an average gain of +11.99~pp over the students trained from scratch, 1.30~pp above the strongest baseline, RSD (+10.69~pp), and is the best logit-based method on every pair, by 0.62 to 6.15~pp. These results are obtained without ground-truth labels ($\lambda=0$), whereas the compared methods use label supervision, so the teacher outputs compared through the contrastive objective suffice to transfer the knowledge of heterogeneous teachers. RKD \citep{relational} and CRD \citep{crd}, which exploit inter-sample relations in feature space, yield average gains of $-1.40$ and $-0.02$~pp, suggesting that relations transfer more readily across architectures when measured in a space that both networks share.
\textbf{CNN students.} HACD ranks first on all six pairs and exceeds RSD, the strongest feature-based method, by 1.08 to 3.26~pp while using only logits. The gains are largest for MobileNetV2: with Swin-T, ViT-S, and Mixer-B/16 teachers, HACD reaches 85.38\%, 84.66\%, and 84.20\%, 3.11 to 3.85~pp above HCD \citep{hcd}. With the Swin-T teacher, the MobileNetV2 student (85.38\%) surpasses the ResNet-18 student (85.00\%), although its scratch accuracy is lower.
\textbf{ViT students.} HACD ranks first on all four pairs. It exceeds RSD by 0.25 to 2.96~pp on three pairs and is on par with it on ConvNeXt-T $\rightarrow$ DeiT-T (82.50\% vs.\ 82.46\%). Against the strongest logit-based method, LDRLD \citep{ldrld}, the margins range from 1.75 to 6.15~pp.
\textbf{MLP students.} The gains of HACD over the scratch students (+15.67 and +15.28~pp) are the largest among all 12 pairs. HACD is the best logit-based method on both pairs, exceeding OFA \citep{ofa} by 1.60 and 0.62~pp, and ranks second with the Swin-T teacher (82.23\%) and third with the ConvNeXt-T teacher (81.84\%) among all methods, after the feature-based RSD and PAT \citep{pat}.

\subsection{Heterogeneous Distillation on ImageNet-1K}
\label{sec:imagenet_results}

Table~\ref{tab:imagenet_results} reports the results on the 15 heterogeneous teacher-student pairs of ImageNet-1K, where label supervision is used ($\lambda=1$, Section~\ref{sec:experimental_setup}). HACD achieves the best accuracy on 12 pairs and the second-best on two. It improves over KD on every pair and attains the highest average gain over the students trained from scratch, +2.89~pp, compared with +2.34~pp for RSD and +2.20~pp for OFA.
\textbf{CNN students.} HACD ranks first on five of the six pairs, exceeding the strongest competitor by 0.15 to 0.81~pp, and second on DeiT-T $\rightarrow$ ResNet-18 (71.50\% vs.\ 71.70\% for RSD). The MobileNetV2 student benefits most: with DeiT-T, Swin-T, and Mixer-B/16 teachers, it reaches 72.88\%, 73.38\%, and 73.13\%, gains of 4.01 to 4.51~pp over its scratch accuracy, the largest among the 15 pairs.
\textbf{ViT students.} HACD ranks first on four of the six pairs, by 0.27 to 1.43~pp. The margins are largest for the DeiT-T student, which reaches 75.89\% with ConvNeXt-T and 75.76\% with Mixer-B/16, 1.43 and 1.30~pp above the strongest competitor. With the ResNet-50 teacher, the strongest results are those of OFA and RSD combined with FitNets (marked $^{*}$); HACD, which uses only the logits, matches RSD on DeiT-T (76.17\%).
\textbf{MLP students.} HACD ranks first on all three pairs, exceeding the strongest competitor by 0.18 to 0.74~pp.

\section{Ablation Study and Analysis}
\label{sec:ablation}

Except for the class temperature, which is analyzed on ImageNet-1K teachers, the experiments in this section are conducted on CIFAR-100 with two pairs that reverse the direction of the architectural mismatch: Swin-T $\rightarrow$ ResNet-18 (Transformer teacher, CNN student) and ConvNeXt-T $\rightarrow$ DeiT-T (CNN teacher, ViT student). Other settings follow Section~\ref{sec:experimental_setup}.

\begin{table}[t]
\centering
\caption{Ablation of the components of HACD on CIFAR-100 (top-1, \%). M/S: moderate/strong view; Student/Teacher: views processed by each network. Row (a): KD result of Table~\ref{tab:cifar100_results}. KD$^{*}$: KD loss ($T=4$) under the training setting of HACD. Row (f) is HACD.}
\label{tab:ablation_components}
\resizebox{\columnwidth}{!}{%
\begin{tabular}{cccc cc}
\toprule
& Objective & Student & Teacher & \shortstack{Swin-T\\$\rightarrow$ ResNet-18} & \shortstack{ConvNeXt-T\\$\rightarrow$ DeiT-T} \\
\midrule
(a) & KD          & --   & --   & 78.74 & 72.99 \\
(b) & KD$^{*}$    & M    & M    & 79.68 & 81.21 \\
(c) & Contrastive & M    & M    & 83.35 & 81.84 \\
(d) & Contrastive & S    & M    & 83.08 & 74.27 \\
\midrule
(e) & Contrastive & M, S & S    & 84.41 & 80.96 \\
\rowcolor{gray!15}
(f) & Contrastive & M, S & M    & \textbf{85.00} & \textbf{82.50} \\
\bottomrule
\end{tabular}%
}
\end{table}

\textbf{Components.} Table~\ref{tab:ablation_components} separates the objective from the views, with one teacher pass per sample in every variant. Under the same view and training configuration, replacing KD$^{*}$ with the contrastive objective improves the accuracy by 3.67 and 0.63~pp (row (b) $\rightarrow$ (c)), and the full HACD improves it by 5.32 and 1.29~pp (row (b) $\rightarrow$ (f)). Unlike KD$^{*}$, which matches each softened distribution in isolation, the contrastive objective also requires each student output to be closer to its own target than to the targets of the other samples. As the sole student input, the strong view is 0.27 and 7.57~pp below the moderate view (row (c) $\rightarrow$ (d)), whereas as a second student view it adds 1.65 and 0.66~pp (row (c) $\rightarrow$ (f)). Using strong-view teacher targets as the only supervision (row (e)) lowers the accuracy by 0.59 and 1.54~pp relative to HACD, since each teacher target also shapes the supervision of the other samples in the batch.

\begin{figure}[t]
\centering
\includegraphics[width=\columnwidth]{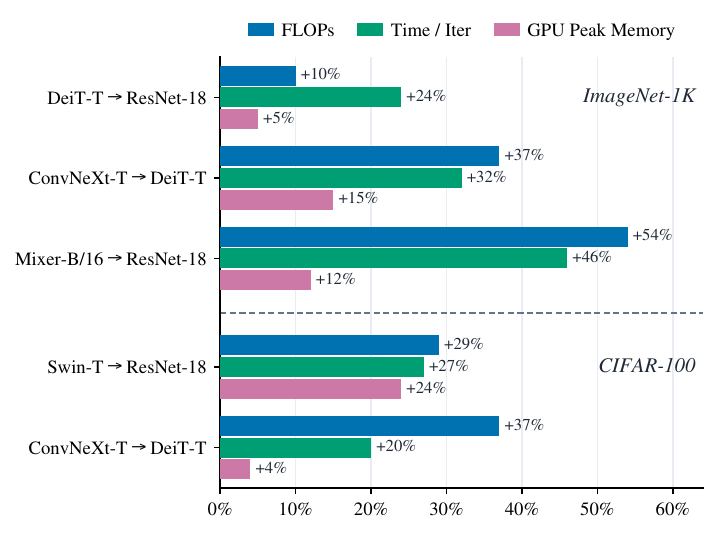}
\caption{Overhead of symmetric teacher views over HACD in training FLOPs, time per iteration, and peak GPU memory.}
\label{fig:efficiency_symmetric}
\end{figure}

\textbf{Symmetric teacher views.} Aligning each student view with the teacher output on the same view requires a second teacher forward pass per sample. This raises the accuracy from 85.00\% and 82.50\% to 85.21\% and 83.63\%, at a cost that grows with the size of the teacher: across the five pairs in Fig.~\ref{fig:efficiency_symmetric}, the training FLOPs increase by 10\% to 54\%, the time per iteration by 20\% to 46\%, and the peak memory by 4\% to 24\%. With row (e), this result indicates that strong-view targets help as a complement (weight $1-\alpha=0.3$) and hurt as the primary supervision. HACD keeps the moderate view as the only source of targets, with a single teacher pass per sample regardless of the number of student views.

\begin{figure}[t]
\centering
\includegraphics[width=\columnwidth]{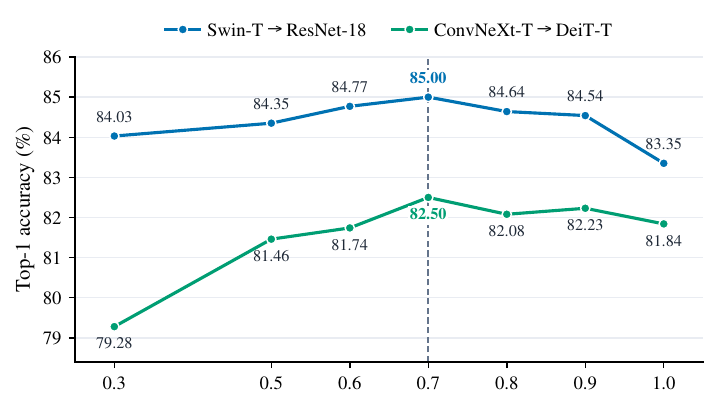}
\caption{Sensitivity to the view weight $\alpha$ on CIFAR-100. The dashed line marks the default $\alpha=0.7$.}
\label{fig:ablation_alpha}
\end{figure}

\textbf{Hyperparameters.} HACD uses the same $\alpha$, $\tau$, and $\lambda$ for all pairs of a dataset, without per-pair tuning. Fig.~\ref{fig:ablation_alpha} varies the view weight $\alpha$. The default $\alpha=0.7$ is the most accurate setting on both pairs, and for $\alpha\in[0.6,0.9]$ the accuracy stays within 0.46 and 0.76~pp of it. Shifting the weight towards the strong view lowers the accuracy by 0.65 and 1.04~pp at $\alpha=0.5$ and by 0.97 and 3.22~pp at $\alpha=0.3$, while removing the strong view ($\alpha=1$) lowers it by 1.65 and 0.66~pp. The default $\tau=0.2$ is also the best on both pairs (Table~\ref{tab:ablation_tau_lambda}). A lower $\tau=0.1$ concentrates the loss on the hardest negatives and costs 0.77 and 1.56~pp, whereas a higher $\tau=0.5$ weakens the separation between samples with similar teacher outputs and costs 0.61 and 1.13~pp. Ground-truth labels affect the two students in opposite directions: $\lambda=0.5$ lowers the accuracy of ResNet-18 by 1.07~pp but raises that of DeiT-T by 0.68~pp, and $\lambda=1.0$ lowers both, by 1.77 and 0.91~pp.

\begin{table}[t]
\centering
\caption{Sensitivity to $\tau$ and $\lambda$ on CIFAR-100 (top-1, \%). Defaults are shaded.}
\label{tab:ablation_tau_lambda}
\resizebox{\columnwidth}{!}{%
\begin{tabular}{l ccc ccc}
\toprule
& \multicolumn{3}{c}{$\tau$} & \multicolumn{3}{c}{$\lambda$} \\
\cmidrule(lr){2-4}\cmidrule(lr){5-7}
Pair & 0.1 & \cellcolor{gray!15}0.2 & 0.5 & \cellcolor{gray!15}0 & 0.5 & 1.0 \\
\midrule
Swin-T $\rightarrow$ ResNet-18  & 84.23 & \cellcolor{gray!15}\textbf{85.00} & 84.39 & \cellcolor{gray!15}\textbf{85.00} & 83.93 & 83.23 \\
ConvNeXt-T $\rightarrow$ DeiT-T & 80.94 & \cellcolor{gray!15}\textbf{82.50} & 81.37 & \cellcolor{gray!15}82.50 & \textbf{83.18} & 81.59 \\
\bottomrule
\end{tabular}%
}
\end{table}

\begin{figure}[t]
\centering
\includegraphics[width=\columnwidth]{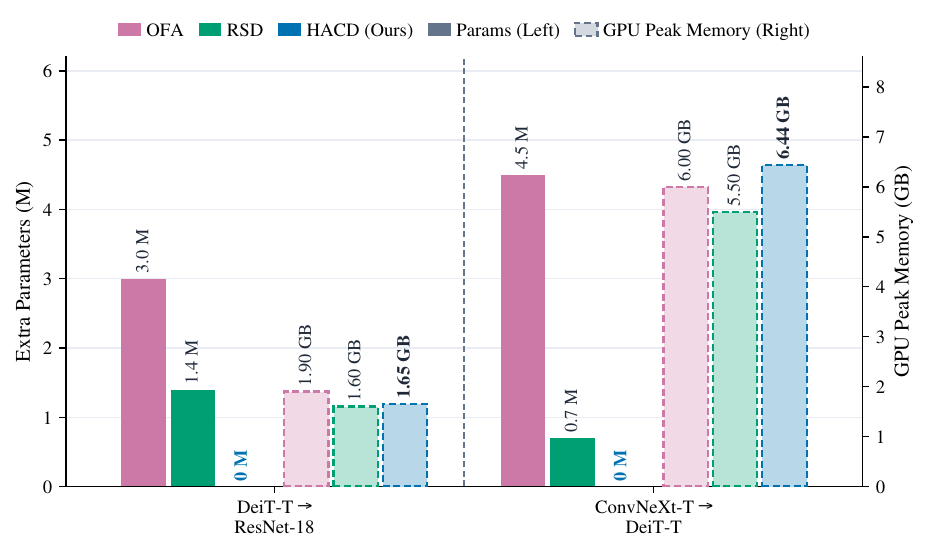}
\caption{Extra learnable parameters (solid, left axis) and peak GPU memory per training step (dashed, right axis) on ImageNet-1K.}
\label{fig:efficiency_methods}
\end{figure}

\textbf{Training cost.} HACD introduces no learnable parameters, memory bank, or momentum encoder, whereas OFA and RSD train 3.0 and 1.4~M extra parameters for DeiT-T $\rightarrow$ ResNet-18 and 4.5 and 0.7~M for ConvNeXt-T $\rightarrow$ DeiT-T (Fig.~\ref{fig:efficiency_methods}). The peak memory of HACD, 1.65 and 6.44~GB, is comparable to that of OFA (1.90 and 6.00~GB) and RSD (1.60 and 5.50~GB). Its overhead comes from the student activations rather than from alignment modules: the second student view, which yields the gain of Table~\ref{tab:ablation_components}, row (c) $\rightarrow$ (f), accounts for 0.65 and 2.15~GB.

\begin{figure}[t]
\centering
\begin{subfigure}[t]{0.49\columnwidth}
\centering
\includegraphics[width=\linewidth]{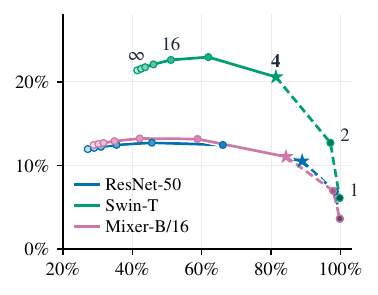}
\caption{Tail removed vs.\ dark knowledge kept}
\label{fig:temperature_knowledge}
\end{subfigure}\hfill
\begin{subfigure}[t]{0.49\columnwidth}
\centering
\includegraphics[width=\linewidth]{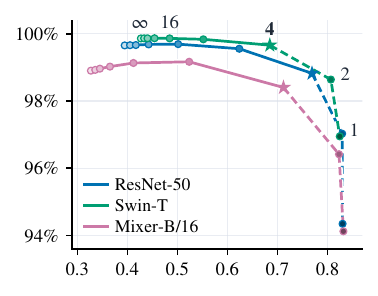}
\caption{Class structure vs.\ instance discrimination}
\label{fig:temperature_tradeoff}
\end{subfigure}
\caption{Representation $u_T$ of three ImageNet-1K teachers as $T$ decreases from $\infty$ to 1 (solid: $T\geq4$; dashed: $T<4$; stars: $T=4$). (a) Energy in the ten highest-ranked classes (horizontal) versus energy in ranks 2--10 (vertical). (b) Class separation (horizontal) versus instance discrimination (vertical).}
\label{fig:temperature}
\end{figure}

\textbf{Class temperature.} The class temperature sets the space in which the relations are compared (Section~\ref{sec:logit_representation}). Fig.~\ref{fig:temperature} follows the representation of a CNN, a Transformer, and an MLP teacher on ImageNet-1K as $T$ decreases. In the high-temperature limit, the ten highest-ranked classes hold only 27.1\%--41.4\% of the energy, so the long tail dominates. At $T=4$, this share rises to 81.4\%--88.9\%, while ranks 2--10, which carry the dark knowledge, retain 10.5\%--20.6\% (Fig.~\ref{fig:temperature_knowledge}); the class separation increases from 0.33--0.43 to 0.69--0.77, and instance discrimination remains at 98.4\%--99.7\% (Fig.~\ref{fig:temperature_tradeoff}). Lower temperatures remove the dark knowledge and reduce instance discrimination, consistent with Propositions~\ref{prop:limits} and~\ref{prop:support}. We therefore use $T=4$ for ImageNet-1K teachers, where it suppresses the long tail and keeps the dark knowledge and instance discrimination. CIFAR-100 teachers already place 84\% of the energy in the ten highest-ranked classes, so the centered logit direction is used directly.

\section{Conclusion}
\label{sec:conclusion}
We presented HACD, a logit-only method for heterogeneous knowledge distillation that transfers the relational structure of teacher predictions instead of matching their architecture-specific values directly. It compares teacher and student outputs as centered, temperature-softened probability directions, transfers their relations through bidirectional instance-level contrast, and computes the teacher targets on a moderate view while the student also learns from a strong view. Under an identical training configuration, the contrastive objective outperforms per-sample matching, and moderate-view teacher targets give the most accurate design at one teacher pass per sample. Without feature alignment modules or extra learnable parameters, HACD achieves the best or second-best accuracy on 25 of the 27 heterogeneous pairs across CNNs, Transformers, and MLPs in comparison with published results, and on CIFAR-100 it needs no ground-truth labels in distillation to do so. These results suggest that, across architectural families, transferring how a teacher relates samples to one another complements, and can outperform, matching the values it assigns to each sample. Because HACD needs only a shared label space, possible extensions include selecting $T$ from the teacher output distribution, dense prediction with per-pixel or per-region logits, whose relations could be formed among spatial positions as well as images, and distillation across different label spaces.

\appendix
\section{Theoretical Analysis of the Representation}
\label{app:theory}

This appendix proves the properties of the representation stated in Section~\ref{sec:logit_representation}. Throughout, let $p=p^{T}(z)$, $\mathbf{d}=\Pi p$, and $N_{\mathrm{eff}}=\big(\sum_{c=1}^{C}p_c^{2}\big)^{-1}$, the effective number of classes, and let $\mathbf{e}_k$ denote the $k$-th standard basis vector. We assume $z\notin\operatorname{span}\{\mathbf{1}\}$, so that $p$ is not uniform and $u_T(z)=\mathbf{d}/\|\mathbf{d}\|_2$ is well defined. Since $\mathbf{1}^{\top}p=1$,
\begin{equation}
\begin{gathered}
    p=\tfrac{1}{C}\mathbf{1}+\mathbf{d},\qquad
    \mathbf{1}^{\top}\mathbf{d}=0,\\
    \|p\|_2^{2}=\tfrac{1}{C}+\|\mathbf{d}\|_2^{2}=N_{\mathrm{eff}}^{-1}.
\end{gathered}
\label{eq:app_decomposition}
\end{equation}

\subsection{Offset Invariance and Necessity of Centering}
\label{app:centering}

\begin{proposition}
\label{prop:centering}
(a) $u_T(z+c\mathbf{1})=u_T(z)$ for all $c\in\mathbb{R}$.
(b) Without centering, i.e., with $p^{T}(z)/\|p^{T}(z)\|_2$ as the representation, any two inputs whose softened predictions satisfy $N_{\mathrm{eff}}\ge C/(1+\epsilon)$ with $\epsilon\ge0$ have
\begin{equation*}
    \cos\!\big(p^{T}(z_i),\,p^{T}(z_j)\big)\;\ge\;\frac{1-\epsilon}{1+\epsilon}.
\end{equation*}
\end{proposition}

\begin{proof}
(a) The softmax is invariant to adding a constant to all logits, since the factor $e^{c/T}$ cancels between its numerator and denominator; hence $p^{T}(z+c\mathbf{1})=p^{T}(z)$.

(b) By Eq.~\eqref{eq:app_decomposition}, the condition on $N_{\mathrm{eff}}$ is equivalent to $\|\mathbf{d}_i\|_2^{2}=N_{\mathrm{eff},i}^{-1}-\tfrac{1}{C}\le\epsilon/C$, and likewise for $j$. Using Eq.~\eqref{eq:app_decomposition} and the Cauchy--Schwarz inequality,
\begin{align*}
    p_i^{\top}p_j&=\tfrac{1}{C}+\mathbf{d}_i^{\top}\mathbf{d}_j\ge\tfrac{1}{C}-\|\mathbf{d}_i\|_2\|\mathbf{d}_j\|_2\ge\frac{1-\epsilon}{C},\\
    \|p_i\|_2\|p_j\|_2&=\sqrt{\tfrac{1}{C}+\|\mathbf{d}_i\|_2^{2}}\,\sqrt{\tfrac{1}{C}+\|\mathbf{d}_j\|_2^{2}}\le\frac{1+\epsilon}{C}.
\end{align*}
For $\epsilon<1$, dividing the two bounds yields the claim. For $\epsilon\ge1$, the bound is non-positive, whereas the cosine is non-negative because $p_i$ and $p_j$ have non-negative entries.
\end{proof}

Proposition~\ref{prop:centering}(b) shows that uncentered representations collapse whenever the softened predictions are close to uniform, which is common for a large number of classes and a moderate temperature. All similarities in Eq.~\eqref{eq:contrastive_similarity} then approach the same value, and Eq.~\eqref{eq:bidirectional_contrastive} degenerates to its chance level $\log B$. The cause is the term $\tfrac{1}{C}$ in $p_i^{\top}p_j$, which all samples share; the projection $\Pi$ removes exactly this component.

\subsection{Temperature Limits}
\label{app:limits}

\begin{proposition}
\label{prop:limits}
Let $\tilde z=\Pi z\neq 0$.
(a) As $T\to\infty$,
\begin{equation*}
    \Pi\,p^{T}(z)=\frac{1}{CT}\Big[\tilde z+\frac{1}{2T}\,\Pi\big(\tilde z\odot\tilde z\big)\Big]+O\!\left(T^{-3}\right),
\end{equation*}
and hence $u_T(z)\to\tilde z/\|\tilde z\|_2$, the centered logit direction.
(b) As $T\to0$, if $\hat c=\arg\max_c z_c$ is unique,
\begin{equation*}
    u_T(z)\to\frac{\mathbf{e}_{\hat c}-\tfrac{1}{C}\mathbf{1}}{\sqrt{1-1/C}}.
\end{equation*}
\end{proposition}

\begin{proof}
(a) By Proposition~\ref{prop:centering}(a), $p^{T}(z)=p^{T}(\tilde z)$. Let $m_2=\tfrac{1}{C}\sum_c\tilde z_c^{2}$. Expanding the exponential to second order and using $\sum_c\tilde z_c=0$,
\begin{align*}
    e^{\tilde z_c/T}&=1+\frac{\tilde z_c}{T}+\frac{\tilde z_c^{2}}{2T^{2}}+O\!\left(T^{-3}\right),\\
    \sum_{k}e^{\tilde z_k/T}&=C\Big(1+\frac{m_2}{2T^{2}}\Big)+O\!\left(T^{-3}\right).
\end{align*}
Dividing the two expansions gives
\begin{equation*}
    p_c=\frac{1}{C}\Big[1+\frac{\tilde z_c}{T}+\frac{\tilde z_c^{2}-m_2}{2T^{2}}\Big]+O\!\left(T^{-3}\right).
\end{equation*}
Subtracting $\tfrac{1}{C}\mathbf{1}$ and using $\tilde z\odot\tilde z-m_2\mathbf{1}=\Pi(\tilde z\odot\tilde z)$ yields the expansion. Hence $CT\,\Pi p=\tilde z+O(T^{-1})\to\tilde z\neq0$, and since normalization is continuous away from zero, $u_T(z)\to\tilde z/\|\tilde z\|_2$.

(b) Dividing the numerator and the denominator of the softmax by $e^{z_{\hat c}/T}$,
\begin{equation*}
    p_c=\frac{\exp\!\big((z_c-z_{\hat c})/T\big)}{1+\sum_{k\neq\hat c}\exp\!\big((z_k-z_{\hat c})/T\big)}.
\end{equation*}
Since $z_k<z_{\hat c}$ for all $k\neq\hat c$, $p\to\mathbf{e}_{\hat c}$ as $T\to0$, and therefore $\Pi p\to\mathbf{e}_{\hat c}-\tfrac{1}{C}\mathbf{1}$. Its squared norm is $(1-\tfrac{1}{C})^{2}+\tfrac{C-1}{C^{2}}=1-\tfrac{1}{C}>0$, and normalization yields the limit.
\end{proof}

Proposition~\ref{prop:limits}(a) shows that the centered logit direction is a special case of Eq.~\eqref{eq:logit_normalization}, which therefore strictly generalizes it. Proposition~\ref{prop:limits}(b) shows that, at low temperature, two samples whose teacher outputs share the same predicted class obtain the same representation, $k_j\to k_i$. Since $|S_{ij}-S_{ii}|\le\|k_j-k_i\|_2/\tau\to0$, every such negative becomes indistinguishable from the positive, and the instance-level contrast degenerates into a prediction-level contrast.

\subsection{Monotone Class Support}
\label{app:support}

\begin{proposition}
\label{prop:support}
$N_{\mathrm{eff}}$ is non-decreasing in $T$, with $N_{\mathrm{eff}}\to1$ as $T\to0$ (for a unique maximum) and $N_{\mathrm{eff}}\to C$ as $T\to\infty$. Moreover, for every class $k$,
\begin{equation*}
    \frac{\partial u_T(z)}{\partial z_k}=\frac{p^{T}_{k}}{T\,\|\Pi\,p^{T}\|_2}\,\big(I-u_Tu_T^{\top}\big)\,\Pi\,\big(\mathbf{e}_k-p^{T}\big),
\end{equation*}
and consequently
\begin{equation*}
    \Big\|\frac{\partial u_T(z)}{\partial z_k}\Big\|_2\le\frac{\sqrt{2}\,p^{T}_{k}}{T\,\|\Pi\,p^{T}\|_2},
\end{equation*}
so the sensitivity of the representation to class $k$ scales with $p^{T}_{k}$.
\end{proposition}

\begin{proof}
Let $\beta=1/T$, $Z(\beta)=\sum_c e^{\beta z_c}$, and let $\mathbb{E}_{\beta}[\cdot]$ and $\mathrm{Var}_{\beta}(\cdot)$ denote the mean and variance of $z_c$ under $c\sim\operatorname{softmax}(\beta z)$. Then $\tfrac{d}{d\beta}\log Z(\beta)=\mathbb{E}_{\beta}[z]$ and $\tfrac{d}{d\beta}\mathbb{E}_{\beta}[z]=\mathrm{Var}_{\beta}(z)\ge0$. Since $\sum_c p_c^{2}=Z(2\beta)/Z(\beta)^{2}$,
\begin{equation*}
    \frac{d}{d\beta}\log\sum_{c}p_c^{2}=2\big(\mathbb{E}_{2\beta}[z]-\mathbb{E}_{\beta}[z]\big)\ge0;
\end{equation*}
hence $N_{\mathrm{eff}}$ is non-decreasing in $T$. The two limits follow from $p\to\mathbf{e}_{\hat c}$ as $T\to0$ and $p\to\tfrac{1}{C}\mathbf{1}$ as $T\to\infty$. For the Jacobian, $\partial p/\partial z_k=\tfrac{p_k}{T}(\mathbf{e}_k-p)$, and differentiating $u_T=\Pi p/\|\Pi p\|_2$ gives $\partial u_T/\partial z_k=\tfrac{1}{\|\Pi p\|_2}\big(I-u_Tu_T^{\top}\big)\Pi\,\partial p/\partial z_k$, which yields the stated expression. Both $I-u_Tu_T^{\top}$ and $\Pi$ are orthogonal projections, and $\|\mathbf{e}_k-p\|_2^{2}=(1-p_k)^{2}+\sum_{c\neq k}p_c^{2}\le2(1-p_k)^{2}\le2$, which gives the bound.
\end{proof}

\bibliographystyle{elsarticle-num-names}
\bibliography{main}

\end{document}